RESEARCH ARTICLE                                                                OPEN ACCESS

# Brain Tumor Detection Based on Bilateral Symmetry Information

Narkhede Sachin, Dr. Deven Shah, Prof. Vaishali khairnar, Prof. Sujata Kadu
G (Student ME (IT) Mumbai University)
 (Terna COE Mumbai),
 (HOD (IT) Terna COE Mumbai)
(Terna COE Mumbai)

**Abstract**
Advances in computing technology have allowed researchers across many fields of endeavor to collect and maintain vast amounts of observational statistical data such as clinical data, biological patient data, data regarding access of web sites , financial data, and the like.
Brain Magnetic Resonance Imaging (MRI) segmentation is a complex problem in the field of medical imaging despite various presented methods. MR image of human brain can be divided into several sub-regions especially soft tissues such as gray matter, white matter and cerebrospinal fluid. Although edge information is the main clue in image segmentation, it can't get a better result in analysis the content of images without combining other information. The segmentation of brain tissue in the magnetic resonance imaging (MRI) is very important for detecting the existence and outlines of tumors. In this thesis , an algorithm about segmentation based on the symmetry character of brain MRI image is presented. Our goal is to detect the position and boundary of tumors automatically. Experiments were conducted on real pictures, and the results show that the algorithm is flexible and convenient.

## I. Introduction

Image segmentation is used to separate an image into several "meaningful" parts. It is an old research topic, which started around 1970, but there is still no robust solution toward it. There are two main reasons; the first is that the content variety of images is too large, and the second one is that there is no benchmark standard to judge the performance. Image segmentation is identification of homogeneous regions in the image. Many algorithms have been elaborated for gray scale images. However, the problem of segmentation for color images, which convey much more information about objects in scenes, has received much less attention of scientific community. While several surveys of monochrome image segmentation techniques were published, similar surveys for color images did not emerge [1].
Image segmentation is one of the primary steps in image analysis for object identification. The main aim is to recognize homogeneous regions within an image as distinct and belonging to different objects. Segmentation stage does not worry about the identity of the objects. They can be labeled later. The segmentation process can be based on finding the maximum homogeneity in grey levels within the regions identified [2].

Segmentation subdivides an image into its regions of components or objects and an important tool in medical image processing. As an initial step segmentation can be used for visualization and compression. Through identifying all pixels (for two dimensional image) or voxels (for three dimensional image) belonging to an object, segmentation of that particular object is achieved. In medical imaging, segmentation is vital for feature extraction, image measurements and image display [3]. Segmentation of the brain structure from MRI has received paramount importance as MRI distinguishes itself from other modalities and MRI can be applied in the volumetric analysis of brain tissues such as multiple sclerosis, schizophrenia, epilepsy, Parkinson's disease, Alzheimer's disease, cerebral atrophy, etc.

Other important aspect of the segmentation method is the color space from which color features are computed (for instance RGB space with Euclidean color distance). Each segmentation technique is usually based on some mathematical model (theory) and/or algorithmic approach (for instance fuzzy clustering, Markov random field, recursive procedure, bottom-up algorithm etc.). Most of segmentation techniques assume something about the scene which is seen in the image (for instance objects are polyhedral made of dielectric materials).This is an additional knowledge attribute of the given segmentation method [4].

Image segmentation is a process of pixel classification. An image is segmented into subsets by assigning individual pixels to classes. It is an important step towards pattern detection and recognition. Segmentation is one of the first steps in image analysis. It refers to the process of partitioning a digital image into multiple regions (sets of pixels). Each of the pixels in a region is similar with respect to some characteristic or computed property, such as color, intensity, or texture. The level of segmentation is decided by the particular characteristics of the problem being considered. Image segmentation could be further used for object matching between two images. An object of interest is specified in the first image by using the segmentation result of that image; then the specified object is matched in the second image by using the segmentation result of that image [5].

## II. Literature Survey

Automatic image segmentation techniques can be classified into four categories, namely, (1) Clustering Methods, (2) Thresholding Methods, (3) Edge-Detection Methods, and (4) Region-Based Methods [6].

### 1. Clustering Methods
Clustering is a process whereby a data set (pixels) is replaced by cluster; pixels may belong together because of the same color, texture etc. There are two natural algorithms for clustering: divisive clustering and agglomerative clustering. The difficulty in using either of the methods directly is that there are lots of pixels in an image. Also, the methods are not explicit about the objective function that is being optimized. An alternative approach is to write down an objective function and then build an algorithm. The K-means algorithm is an iterative technique that is used to partition an image into K clusters, where each pixel in the image is assigned to the cluster that minimizes the variance between the pixel and the cluster center and is based on pixel color, intensity, texture, and location, or a weighted combination of these factors. This algorithm is guaranteed to converge, but it may not return the optimal solution. The quality of the solution depends on the initial set of clusters and the value of K [7].

### 2. Thresholding Methods
Thresholding is the operation of converting a multilevel image into a binary image i.e., it assigns the value of 0 (background) or 1 (objects or foreground) to each pixel of an image based on a comparison with some threshold value T (intensity or color value). When T is constant, the approach is called global thresholding; otherwise, it is called local thresholding. Global thresholding methods can fail when the background illumination is uneven. Multiple thresholds are used to compensate for uneven illumination. Threshold selection is typically done interactively.

### 3. Edge Detection Methods
Edge detection methods locate the pixels in the image that correspond to the edges of the objects seen in the image. The result is a binary image with the detected edge pixels. Common algorithms used are Sobel, Prewitt, Robert, Canny and Laplacian operators. These algorithms are suitable for images that are simple and noise free; and will often produce missing edges, or extra edges on complex and noisy images.

### 4. Region-Based Methods
The goal of region-based segmentation is to use image characteristics to map individual pixels in an input image to sets of pixels called regions that might correspond to an object or a meaningful part of one. The various techniques are: Local techniques, Global techniques and Splitting and merging techniques. The effectiveness of region growing algorithms depends on the application area and the input image. If the image is sufficiently simple, simple local techniques can be effective. However, on difficult scenes, even the most sophisticated techniques may not produce a satisfactory segmentation.

Edge-based techniques are based on the assumption that pixel values change rapidly at the edge between two regions Operators such as Sobel or Roberts operators can be used to detect the edges. And some post procedures such as edge tracking, gap filling can be used to generate closed curves. Region-based techniques are based on the assumption that adjacent pixels in the same region should be consistent in some properties. Namely, they may have similar characteristic such as grey value, color value or texture. The deformable models are based on curves or surfaces defined within an image that moves due to the influence of certain forces [8]. And the global optimization approaches use a global criterion when segmenting the image.

## III. Problem Statement
Image segmentation is a key step from the image processing to image analysis, it occupy an important place. On the other hand, as the image segmentation, the target expression based on segmentation, the feature extraction and parameter measurement that converts the original image to more abstract and more compact form, it is possible to make high-level image analysis and understanding.

If the input brain image is colorized , it is converted into gray image. First read the red, blue and green value of each pixel and then after formulation, three different values are converted into gray value. The automated edge detection technique is proposed to detect the edges of the regions of interest on the digital images automatically. The method is employed to segment an image into two symmetric regions based on finding pixels that are of similar in nature. The more symmetrical the two regions have, the more the edges are weakened. At the same time, the edges not symmetrical are enhanced. In the end, according to the enhancing effect, the unsymmetrical regions can be detected, which is caused by brain tumor.

## IV. The Proposed Mechanism

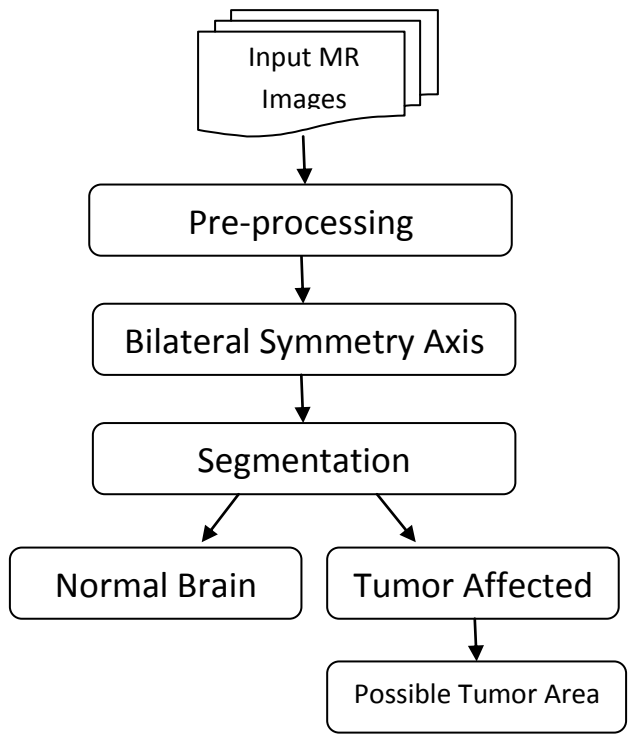

Figure 4.1: Proposed Model

It is based on the image segmentation method, which refers to the major step in image processing, the inputs are images and, outputs are the attributes extracted from those images. It will help to find out symmetric extraction of the brain image.

### 4.1 Data Pre-processing
The MRI images are subject to various types of noises such as irregularities etc. These noises may degrade the quality of the MR image and consequently it cannot provide correct information for subsequent image segmentation and edge detection. In order to improve the quality of the MR image, operations need to be performed to remove or decrease degradations suffered in its acquisition. Preprocessing is also needed in order to homogenize and separate the intensity distributions of the malignant and benign tissues. This can be achieved by using several denoising techniques, viz., Gaussian filter, median filter.

- **Image Smoothing**
Image smoothing act as the pre-processing step for image segmentation, as, almost all of the images suffers from the problem of noise effects. So, pre-processing act as an important aid to the every already existing segmentation methods, in which specialized filters as described above smooth the image and simplifying it for subsequent segmentation step.

- **Image Contrast Enhancement**
Poor contrast is usually one of the most common defects found in the acquired image. This degradation probably is caused by inadequate aperture size and noise. Sometimes this is caused of non linear mapping of the image intensity. The effect of such defects has a great impact on the contrast of the acquired image. In this case, the gray level of each pixel is scaled to improve the contrast of the acquired image. Contrast enhancement step sometimes proves to be one of the important pre-processing steps, especially in case when image has a poor contrast. In the present work, the contrast of the smoothened image is enhanced using the image processing toolbox functions. This improves the visualization of the original image and thus makes the object of interest more clearly visible.

In the first step proper threshold is chosen in order to distinguish the interior area from other organs in the MR image dataset. Then gradient magnitude is computed by using one of Robert, Prewitt or Canny operator and employed as the definition of homogeneity criterion. This implementation allowed stable boundary detection when the gradient suffers from intersection variations and gaps. By analyzing the gradient magnitude, the sufficient contrast present on the boundary region that increases the accuracy of segmentation.

### 4.2 Bilateral Symmetry Axis
Bilateral symmetry axis defining is a straightforward evaluation method that is commonly used for comparing the corresponding MRI brain images to determine the ROI in the image.

Structural and functional asymmetry in the human brain and nervous system is reviewed in a historical perspective. Brain asymmetry is one of such examples, which is a difference in size or shape, or both. Asymmetry analysis of brain has great importance because it is not only indicator for brain cancer but also predict future potential risk for the same. In our work, we have concentrated to segment the anatomical regions of brain, isolate the border line of each to investigate the presence of asymmetry of anatomical regions in MRI. The term asymmetry is often substituted for the term laterality when it comes to left–right differences in psychology and the neurosciences. However, while the term asymmetry can mean both structural and functional left–right dissimilarities, laterality is typically only used in relation to functional asymmetry.

### 4.3 Segmentation

The goal of segmentation is to isolate the regions of interest depending on the problem and its characters. A gray level image consists of two main features, namely region and edge. Segmentation algorithms for gray images are generally based on of two basic properties of image intensity values, discontinuity and similarity. The method is employed to segment an image into two symmetric regions based on finding pixels that are of similar in nature. The method has the advantage that it is fairly robust, quick, and parameter free except for its dependency on the order of pixel processing. The more symmetrical the two regions have, the more the edges are weakened. At the same time, the edges not symmetrical are enhanced. In the end, according to the enhancing effect, the unsymmetrical regions can be detected, which is caused by brain tumor. The output is shown in Figure 3.3

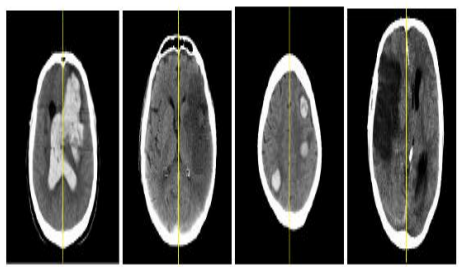

Figure 4.2: Ideal symmetry axis is extracted as the intersection of two halves brain image

### V. Methodology Used

Step 1: Use canny edge detection technique for the finding the edges in brain image.
Step 2: Determine the Mid pixel position of the row and read the intensity of Mid pixel of row
Step 3: Fit the curve by using LSM and Crammer rule.
Step 4: Show the curve in tumor affected area.
Step 5: Calculate and Show the tumor area by using Automatic brain tumor detection

### VI. Performance Analysis

Performance analysis is looking at program execution to pinpoint where bottlenecks or other performance problems might occur. Once you know where potential trouble spots are, you can change your code to remove or reduce their impact. Experimental analysis and statistical analysis are carried out to analyze the performance of the system.

Evaluation of the images showed that under noisy conditions Canny, Prewitt, Robert, exhibit better performance, respectively. Canny yielded the best results as shown in Figure 6.1. This was expected as Canny edge detection accounts for regions in an image. Canny yields thin lines for its edges by using non-maximal suppression. Canny also utilizes hysteresis with thresholding.

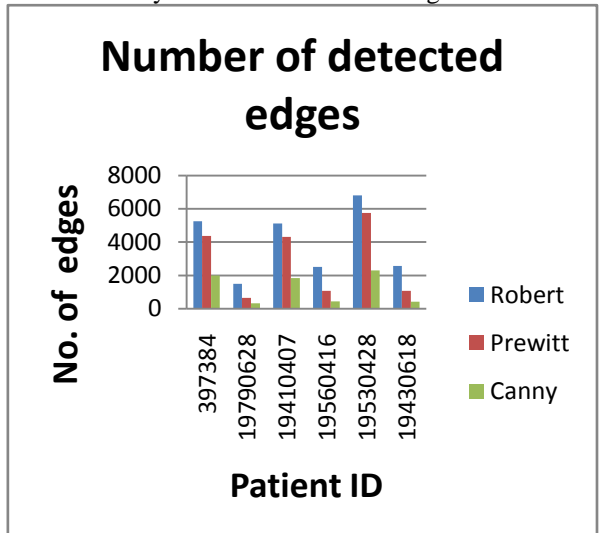

Graph 6.1: Number of detected edges

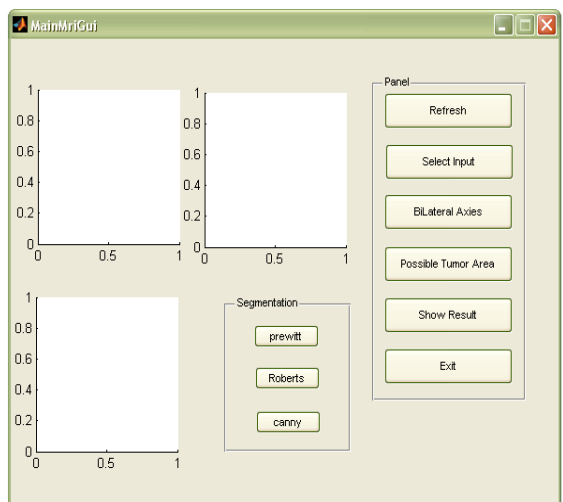

Screen 6.1 Starting window

Click on "Select Input" button then same window will appear contains a grey/color image.

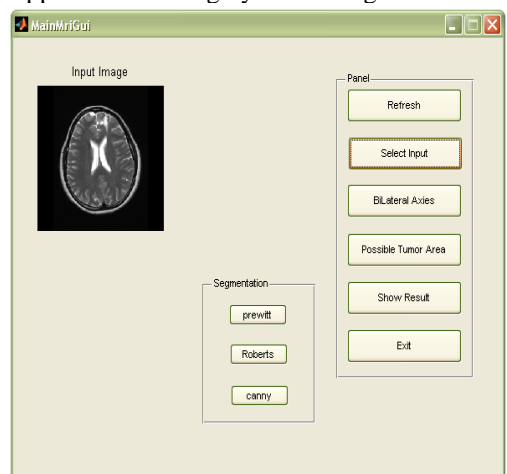

Screen 6.2 : Input Image from database for segmentation

Click on any one of "Segmentation" button & then on "Bilateral Axis" button then figure window will appear which contains number of edges for that particular algorithm and same GUI contains a bilateral axis image.

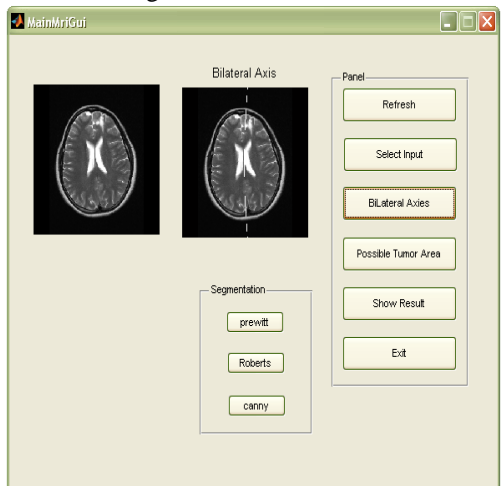

Screen 6.3: Bilateral Axis

In next step, Click on "Possible Tumor Area" button then figure window will appear on same GUI and the detected possible tumor area are shown in red/green/blue color.

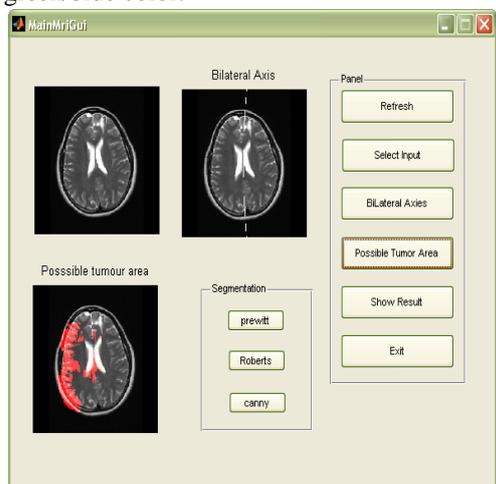

Screen 6.4: Possible Tumor Area

In last step, Click on "Show Result" button then figure window will appear and show volume of possible tumor.

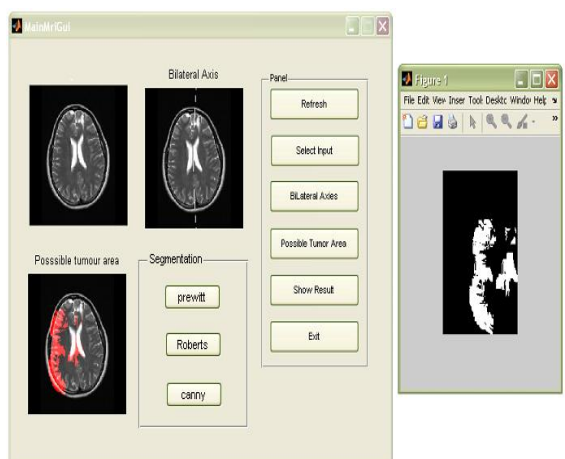

Screen 6.5: Volume of possible tumor area

If tumor is not present, then it will generate message Possible tumor area are not found"

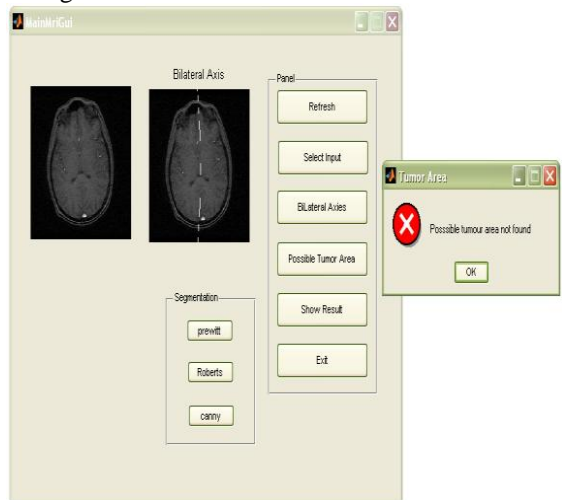

Screen 6.6: Message for healthy brain

### VII. Conclusion

A new system that can be used as a second decision for the surgeons and radiologists is proposed.High grade tumor have more true edges than low grade.MRI of healthy brain has an obviously character almost bilateral symmetrical .However, if there is macroscopic tumor, the symmetry characteristic will be weakened According to the influence on the symmetry by the tumor, we develop a segment algorithm to detect the tumor region automatically